\documentclass{article}

\PassOptionsToPackage{numbers, compress}{natbib}

\usepackage[preprint]{neurips_2026}
\workshoptitle{ML4Molecules: Agentic Systems for Molecular Sciences}

\makeatletter
\renewcommand{\@noticestring}{Accepted to NeurIPS 2026 Workshop on ML4Molecules: Agentic Systems for Molecular Sciences.}
\makeatother

\usepackage[utf8]{inputenc} 
\usepackage[T1]{fontenc}    
\usepackage{hyperref}       
\usepackage{url}            
\usepackage{booktabs}       
\usepackage{amsfonts}   
\usepackage{amsmath}
\usepackage{nicefrac}

\usepackage{graphicx}
\usepackage{float} 
\usepackage{microtype} 
\usepackage{xcolor}
\usepackage{booktabs}
\usepackage{placeins}
\usepackage[table]{xcolor} 
\usepackage{pifont}    
\usepackage{wrapfig}
\usepackage{lipsum}

\newcommand{\cmark}{\ding{51}}%
\newcommand{\xmark}{\ding{55}}

\title{MOFjson: An Explicit Topological Representation of Metal-Organic Frameworks for Language Models}

\author{%
  Anabel Yong      \quad Matthew Hart   \quad Philippe Schwaller \\
  Laboratory of Artificial Chemical Intelligence (LIAC), EPFL \\
  \& National Centre of Competence in Research (NCCR) Separations \\
  \texttt{\{\href{mailto:anabel.yong@epfl.ch}{anabel.yong}, \href{mailto:matthew.hart@epfl.ch}{matthew.hart}, \href{mailto:philippe.schwaller@epfl.ch}{philippe.schwaller}\}@epfl.ch}
}
\begin{document}
\maketitle
\vspace{-15pt}
\begin{abstract}
\vspace{-10pt}
Language models for metal-organic frameworks (MOFs) rely on formats that separate atomic geometry from structural topology, forcing models to choose between unstructured coordinate lists and non-spatial string identifiers\cite{bucior2019identificationmofkeymofid}. We present MOFjson, the first representation separating structural data into a raw atomic layer with explicit bond shift vectors and a derived layer exposing secondary building units (metal nodes and organic linkers) and network topology. Additionally, we introduce a benchmark with crystal-derived ground truth across four question categories. Across four models, MOFjson beats all three baselines in every category: 100\% vs.\ 18--47\% on reaching a block's nearest periodic copy, 96\% vs.\ 71\% on crystal system characterisation, and 60\% vs.\ 48\% on metal node and organic linker coordination. Ablating shift labels drops accuracy to 12.7\%, proving models read the periodic graph rather than the surface format. MOFjson gives language models structure to read rather than reconstruct, establishing a foundation for agentic tool workflows and constrained MOF generation.
\end{abstract}
\vspace{-15pt}
\section{Introduction}
Metal-organic frameworks (MOFs) are porous crystalline materials assembled from metal nodes and organic linkers into periodic networks, with applications in gas storage, carbon capture, and catalysis\cite{ bucior2019identificationmofkeymofid, yaghi2019introductiontoretichem}. The same building blocks can be arranged into different nets with different pore geometries, so a MOF's function follows from its connectivity as much as from its composition \cite{yaghi2019introductiontoretichem}. A representation intended to support property prediction or design must convey both. 

Large language models are increasingly applied to these tasks, and their accuracy depends on how the structure is serialised\cite{runcie2026molecularmoljson, pan2026enhancing}. For small molecules, replacing SMILES or IUPAC strings with an explicit JSON graph schema substantially improves an LLM's ability to translate, reason and generate structures\cite{runcie2026molecularmoljson}. Extending this to MOFs is not a direct transfer. A molecule is a finite graph, whereas a crystal is infinite, and its standard finite encoding, the labelled quotient graph, is one-to-one only once every edge carries an integer vector recording which periodic image it closes onto \cite{xiao2023invertible}. Whether a MOF representation states those labels or leaves them implicit determines what a model can recover from it. Existing MOF representations leave them implicit, and the data-driven MOF domain has been built on them regardless. MOFormer trains a transformer on MOFid strings\cite{cao2023moformer}, MOFGPT generates frameworks autoregressively from MOFid sequences\cite{badrinarayanan2025mofgpt}, and continual pretraining on MOFid alongside CoRE MOF, QMOF, and hMOF improves property prediction when fine-tuning data is scarce\cite{zhao2025core, rosen2021machine, nandy2023database, alrowili2025accelerating}. Generative models rely on it further downstream, using MOFid to decide whether a sampled structure is valid, novel, and unique\cite{simkus2025mofasa}. A descriptor designed for database search now serves as both model input and evaluation criterion. 

We introduce MOFjson, a JSON schema that retains the atom and bond vocabulary of molecular graph schemas and adds two fields addressing the deficiencies above. An integer shift vector on every bond supplies the edge labels the quotient graph requires, so periodic connectivity is stated rather than inferred from coordinates. A second graph over the building blocks states metal node and organic linker assignment and net connectivity directly, so none are recomputed. Labelled quotient graphs\cite{lqg2eon2016topological} are also the natural input for tasks beyond question answering, such as crystal structure prediction, where block-level representations already improve LLM reasoning over MOF assembly\cite{pan2026enhancing, lqg2eon2016topological, lqg-ai2025capturing, lqg3fedorov2017crystal}.

\subsection{Limitations of Existing MOF Representations}
MOFs are distributed as crystallographic information files (CIFs)\cite{cif-hall1991crystallographic}, which record lattice parameters, atomic coordinates, and symmetry operations. A CIF gives geometry but not connectivity: neither which atoms form the metal node or organic linker, nor which bonds close across the cell boundary to complete the net. Both must be re-derived from coordinates, and for cells of hundreds of atoms this is the long, unstructured context current LLMs handle poorly~\cite{pan2026enhancing}. MOFid and MOFkey were built to address ad hoc MOF naming, where one structure appeared under several unrelated names, impeding search and deduplication \cite{bucior2019identificationmofkeymofid}. MOFid encodes a MOF as the SMILES of its building blocks plus a topology code, obtained by running Systre\cite{systredelgado2003identification} on the extracted net and matching it against the Reticular Chemistry Structure Resource (RCSR) database\cite{omaryaghi2008reticularnets}; MOFkey compresses the same information into an InChIKey-style string. As model inputs, they are not sufficiently detailed: a net symbol such as \texttt{tbo} is a pointer into a lookup table, not a statement of how vertices connect, leaving nothing to generalise from beyond the mapping. Coverage is also incomplete, since deconstruction depends on that software converging on the net. Existing representations thus fail in opposite directions: CIF provides exhaustive atomic detail without explicit connectivity, while MOFid and MOFkey offer a compact code with no structural graph left to query. MOFjson labels each bond with the cell it closes into and adds a building-block graph, so connectivity is read rather than inferred and most queries need only that graph, not the atom list.
\section{Methodology}
MOFjson extends MolJSON\cite{runcie2026molecularmoljson}, which stores a finite molecule as typed fields, atoms as nodes and bonds as edges, so a language model reads structure directly instead of decoding a string. Two additions make this work for a crystal. Periodicity: a lattice, and on every bond a label recording which repeat of the cell that bond ends in (Section~\ref{sec:labelledquotientgraph}). Granularity: a coarser graph whose vertices are secondary building units\cite{kalmutzki2018secondarybu}, the metal nodes and organic linkers, so topology is read off rather than recomputed from hundreds of atoms.

The schema separates a lossless record from a computed view. The \texttt{raw} layer is CIF-native: \texttt{lattice}, \texttt{atoms}, and \texttt{bonds} carrying the periodic \texttt{shift} defined below, with optional charge and aromaticity fields; \texttt{roles} labels each atom as a metal node or organic linker. From these, the \texttt{derived} layer caches \texttt{building\_blocks}, the nodes and linkers of MOFid and MOFkey\cite{bucior2019identificationmofkeymofid}; \texttt{net\_edges}, the graph between them; and \texttt{topology}, the matched RCSR symbol \cite{omaryaghi2008reticularnets}. Since \texttt{raw} alone determines the structure, fidelity never rests on the cached layer. Conversion uses \texttt{pymatgen}\cite{pymatgen-ong2013python} and \texttt{ASE}\cite{hjorth2017atomic}; rebuilding a CIF from \texttt{raw} and re-parsing it reproduces the original lattice and connectivity. Table~\ref{tab:representation-comparison} compares representations, and Appendix~\ref{hkust-methodology} shows the representations for the HKUST-1 MOF\cite{hkust-1-Chui1999ACF}.
\begin{table}[h]
\footnotesize
\centering
\caption{Existing MOF formats trade atomic detail for topology; MOFjson preserves both.}
\begin{tabular}{@{}l c c c c@{}}
\toprule
\rowcolor{gray!15}
\textbf{Feature / Property} & \textbf{CIF} & \textbf{MOFid} & \textbf{MOFkey} & \textbf{MOFjson} \\
\midrule
Exact atomic positions & \cmark & \xmark & \xmark & \cmark \\
Explicit bond list & \xmark & \xmark & \xmark & \cmark \\
Node/linker identity & \xmark & \cmark & \cmark & \cmark \\
Net topology symbol & \xmark & \cmark & \cmark & \cmark \\
Boundary-crossing bonds & \xmark & \xmark & \xmark & \cmark{} (\texttt{shift}) \\
Reconstructible to full 3D structure & \cmark{} (native) & \xmark & \xmark & \cmark{} (verified) \\
\bottomrule
\end{tabular}
\label{tab:representation-comparison}
\vspace{-16pt} 
\end{table}
\subsection{Explicit Periodic Graphs in MOFjson}
\label{sec:labelledquotientgraph}
A crystal is one unit cell repeated infinitely along its three lattice vectors, so every atom has a copy in every repeat. Storing a single cell, as a CIF does\cite{batten2012terminology, batten2013mofterminology}, drops which repeat a bond ends in, making a bond that closes onto a neighbouring copy indistinguishable from one that stays inside\cite{xiao2023invertible}. Numbering the repeats by integer coordinates resolves this: each bond $\alpha$ carries a shift $s(\alpha) \in \mathbb{Z}^3$ naming the repeat its second endpoint lies in, $(0,0,0)$ inside the cell and $(1,0,0)$ one repeat along $a$. Labelling edges this way is standard in net construction\cite{boyd2016generalizedgraphs} and dimensionality analysis\cite{gao2020quotientgraph}; what MOFjson does differently is store the labels rather than recompute them per query.
\begin{figure}[H]
    \centering
    \makebox[\textwidth][c]{%
        \includegraphics[trim={0 2.5cm 0 0}, clip, width=0.8\textwidth]{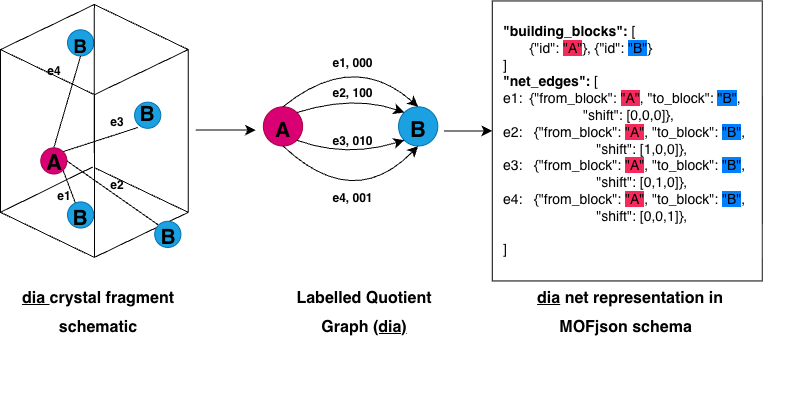}%
    }
    \caption{\textbf{From crystal, to labelled quotient graph, to MOFjson, shown on the \texttt{dia} net.} Net topology abstracts periodic structures into mathematical graphs of nodes and edges to define 3D connectivity\cite{net-topology-def-glasby2024topological}. Here the building blocks are vertices A and B. \textbf{Left:} a crystal fragment, with A bonded to four copies of B in different cells. \textbf{Middle:} the same fragment as a labelled quotient graph, one vertex per building block and one arc per bond with its integer shift vector. \textbf{Right:} MOFjson stores that graph directly, with \texttt{building\_blocks} as vertices and \texttt{net\_edges} as arcs (attachment points omitted). The \texttt{dia} net \cite{dia-blatov2007three} denotes the 4-connected tetrahedral diamond topology.}
\label{fig:mofjson-panel}
    \vspace{-14pt}
\end{figure}
The labels make periodic connectivity computable. Summing shift vectors around a closed walk of bonds $w = (\alpha_1,\dots,\alpha_n)$, negating any bond traversed against its stored direction, yields the net translation vector or \emph{cycle sum} $t(w)$~\cite{boyd2016generalizedgraphs, gao2020quotientgraph}:
\begin{equation*}
t(w) = \sum_{i=1}^{n} \sigma_i\, s(\alpha_i), \qquad \sigma_i = \begin{cases}+1 & \text{walked forward}\\ -1 & \text{walked backward}\end{cases}.
\end{equation*}
A walk with $t(w) = 0$ forms an intra-cell ring, whereas $t(w) \neq 0$ traces a periodic path displaced by $t(w)$; without shifts, representations cannot distinguish the two. MOFjson stores two labelled graphs: the raw atomic layer $G_{\text{atom}}$ and the building-block graph $G_{\text{block}}$, constructed via the map $\pi: V_{\text{atom}} \to V_{\text{block}}$ from connected components in \texttt{roles} (Figure~\ref{fig:mofjson-panel}). Internal bonds are absorbed, while inter-block bonds $\alpha=(u,v)$ become arcs $\pi(\alpha)$ with shifts re-anchored to block frames:
\begin{equation*}
s_{\text{block}}(\pi(\alpha)) = s_{\text{atom}}(\alpha) + o(u) - o(v),
\end{equation*}
where $o(u) \in \mathbb{Z}^3$ is the cell shift offset of the fragment containing atom $u$. These $+o(u) - o(v)$ offsets telescope around any closed walk $w$, ensuring cycle sums over blocks equal cycle sums over atoms so the two layers cannot disagree.

\subsection{Benchmark QnA}
\label{sec:qna}
Four recent systems evaluate MOF question answering. KGQA4MAT scores whether a question is translated into the correct query over a knowledge graph\cite{kg-llm-qna-an2023knowledge}; L$^2$M$^3$OF derives its question-answer pairs from publication abstracts\cite{l2m3ofqna-cui20252m}; MOFReasoner distils reasoning traces from the literature\cite{llm-qna-bai2026mofreasoner}; and MOF-ChemUnity's assistant answers from a graph linking publications to deposited structures, judged by expert preference rather than against a fixed answer key\cite{llmqna-pruyn2025mof}. In each, the answer is recoverable from curated text or metadata, so a model that saw the framework during pretraining can score well without reading the structure it was given. None isolates the representation, and where building-block or topology labels appear, they were generated with MOFid~\cite{kg-llm-qna-an2023knowledge, l2m3ofqna-cui20252m}, so MOFid cannot then be evaluated against them. Our benchmark differs on both counts: the ground truth is computed from the crystal rather than retrieved, and the structure is the input rather than a lookup key, so accuracy measures reading rather than recall (Table~\ref{tab:qna-categories}).

An answer is required on every prompt. A representation lacking the information would otherwise be rewarded for abstaining, and the failure would go unrecorded. Each category targets a descriptor used to specify a MOF. \emph{Symmetry}: the crystal system implied by the space group, computed with \texttt{pymatgen}\cite{pymatgen-ong2013python}, which constrains the pore geometries a framework can adopt\cite{furukawa2010ultrahigh}. \emph{Chemical identity}: which atoms make up a given building block, since a MOF is specified by its node and linker\cite{yaghi2019introductiontoretichem}. \emph{Connectivity}: $\deg(b)$ in $G_{\text{block}}$, the linkers bonded to a block and the parameter that reticular design uses to pick a target net\cite{omaryaghi2008reticularnets}. \emph{Net path}: the fewest bonds forming a closed walk with $t(w) \neq 0$, reaching a different periodic image of the same block.

\begin{table}[H]
\centering
\footnotesize 
\caption{\textbf{One example prompt per QnA category in the benchmark.} \texttt{\{representation\}} marks where the structure data (CIF, MOFjson, MOFid, or MOFkey) is inserted.}
\begin{tabular}{@{}p{0.20\linewidth}p{0.75\linewidth}@{}}
\toprule
\textbf{Category} & \textbf{Example prompt} \\
\midrule
Symmetry & ``In this metal--organic framework \{representation\}, what crystal system does the structure belong to?'' \\
\addlinespace
Chemical identity & ``In this MOF \{representation\}, what metal(s) make up the SBU?'' (or) ``... what is the SMILES representation of the organic linker?'' \\
\addlinespace
Connectivity & ``What is the connectivity of the Co-containing SBU -- how many organic linkers coordinate directly to it in this MOF \{representation\}?'' \\
\addlinespace
Net path & ``In this MOF \{representation\}, how few connections are needed to reach the nearest periodic image of this same Al-containing building unit?'' \\
\bottomrule
\end{tabular}
\label{tab:qna-categories}
\vspace{-14pt} 
\end{table}

\section{Experiments}
\label{sec:experiments}
Each of the 2,737 computation-ready CIFs in the CoRE MOF database\cite{chung2014computation} is converted to MOFjson and back to CIF, with the reconstruction compared against the source with \texttt{pymatgen}'s \texttt{StructureMatcher}\cite{pymatgen-ong2013python}. Only the \texttt{raw} layer enters this test, since the derived fields are recomputable. Every structure is recovered, at cell sizes from under 50 to over 800 atoms (Figure~\ref{fig:conversion-fidelity}, Appendix~\ref{sec:conversion-fidelity}). For the QnA benchmark, performance is reported as the mean $\pm$ standard deviation across 3 independent runs, each drawing 50 random structures that are rendered in four representations (CIF, MOFid, MOFkey, and our MOFjson).

We evaluate the 4 representations across four models: \texttt{gpt-5.6-luna}\cite{openai2026gpt56}, \texttt{claude-haiku-4.5}\cite{anthropic2025haiku}, \texttt{gemini-3.7-flash}\cite{google2026gemini37}, and the open-weight \texttt{apertus-v1.5-70b}\cite{apertus-hernandez2026apertus}. Grading is a fixed comparison against precomputed ground truth: integer match for Connectivity and Net path, case-insensitive string match for Symmetry, and element or SMILES match for Chemical identity. We compare MOFjson against each baseline with a paired Wilcoxon signed-rank test, reporting Cliff's $\delta$\cite{cliff1993dominance}; $p$-values are Holm-Bonferroni corrected across the category-baseline comparisons. 

\section{Results}
\label{sec:results}
\begin{figure}[H]
    \centering
    \makebox[\textwidth][c]{%
        \begin{minipage}[b]{0.60\textwidth}
            \centering
            \includegraphics[width=\textwidth, keepaspectratio]{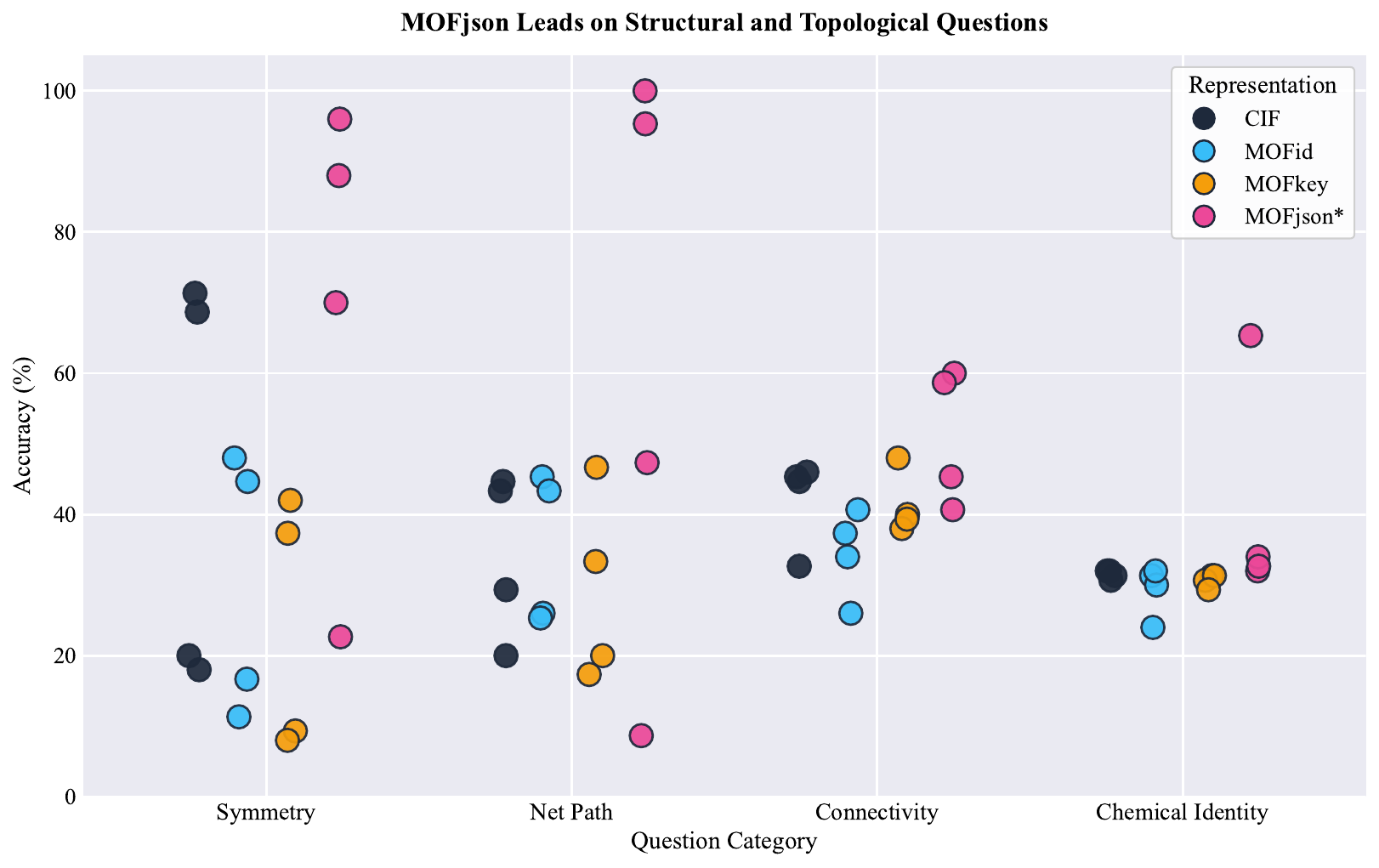}
        \end{minipage}%
        \hfill
        \begin{minipage}[b]{0.60\textwidth}
            \centering
            \includegraphics[width=\textwidth, keepaspectratio]{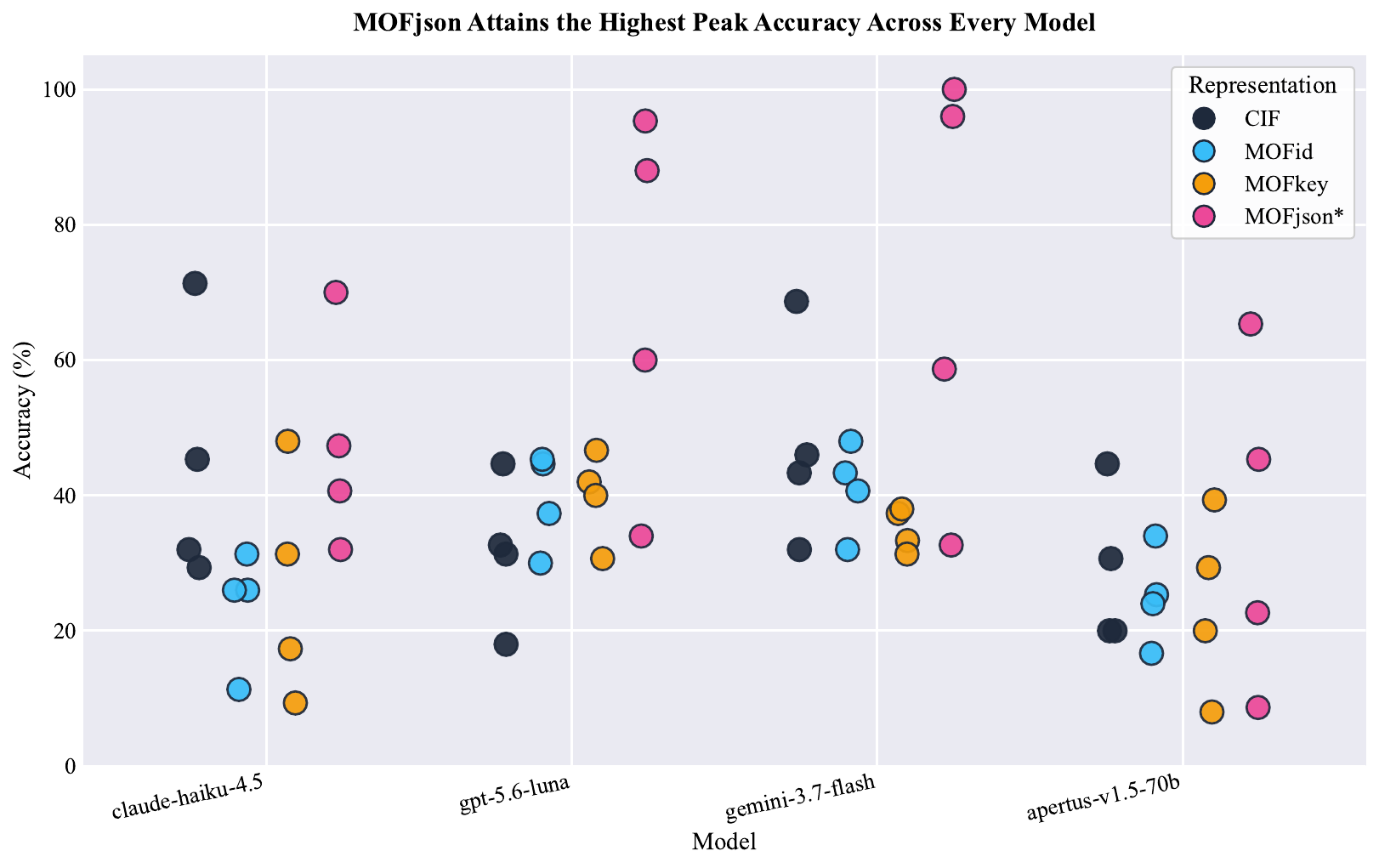}
        \end{minipage}%
    }
    \caption{\textbf{MOFjson yields the highest and most consistent accuracy of any representation tested.} \textbf{Left:} accuracy by question category, each point one model's mean over three runs; MOFjson leads on Symmetry, Net path, and Connectivity, the three requiring structural reasoning, but not Chemical identity, readable from atom labels in any representation. \textbf{Right:} grouped by model; MOFjson gives three of four models their best score. Full per-model, per-representation accuracy is reported in Appendix \ref{sec:breakdownofaccuracy}}
\label{fig:strip-pair}
\end{figure}
MOFjson's advantage is largest where answering requires reading structure out of the representation (Figure~\ref{fig:strip-pair}, left). On Net path it reaches 95--100\% for \texttt{gpt-5.6-luna} and \texttt{gemini-3.7-flash}, against 18--47\% for every baseline. Symmetry follows the same pattern: MOFjson matches or exceeds CIF for three models, while MOFid and MOFkey trail both throughout, consistent with discarding lattice geometry. The advantage narrows where the answer does not depend on topology. Connectivity is mixed, with MOFjson highest overall but behind MOFkey for \texttt{claude-haiku-4.5}, and on Chemical identity all four representations cluster within a few points for three of four models.

Grouped by model (Figure~\ref{fig:strip-pair}, right), the ordering holds for \texttt{gpt-5.6-luna} and \texttt{gemini-3.7-flash} and breaks in two places. \texttt{claude-haiku-4.5} attains its own best score with CIF on Symmetry rather than MOFjson (Table~\ref{tab:model-winners}, Appendix~\ref{tab:best-model-rep}), and \texttt{apertus-v1.5-70b} sits lowest throughout, its MOFjson Net path score falling below all three baselines. Table~\ref{tab:category-winners} reports the best (representation, model) pair for each category: MOFjson wins every one, though its Chemical identity win comes from \texttt{apertus-v1.5-70b}, the only model on which that category separates at all.

\begin{table}[H]
\centering
\footnotesize 
\caption{\textbf{Best-performing (representation, model) pair for each question category.}}
\begin{tabular}{lllr}
\toprule
\textbf{Question Category} & \textbf{Best Representation} & \textbf{Best Model} & \textbf{Accuracy (\% $\pm$ SD)} \\
\midrule
Symmetry & MOFjson & \texttt{gemini-3.7-flash} & 96.0 $\pm$ 4.0 \\
Net path & MOFjson & \texttt{gemini-3.7-flash} & 100.0 $\pm$ 0.0 \\
Connectivity & MOFjson & \texttt{gpt-5.6-luna} & 60.0 $\pm$ 4.0 \\
Chemical identity & MOFjson & \texttt{apertus-v1.5-70b} & 65.3 $\pm$ 8.1 \\
\bottomrule
\end{tabular}
\label{tab:category-winners}
\vspace{-12pt} 
\end{table}
\begin{wrapfigure}{r}{0.4\textwidth}
\centering
\vspace{-10pt}
\includegraphics[width=0.38\textwidth]{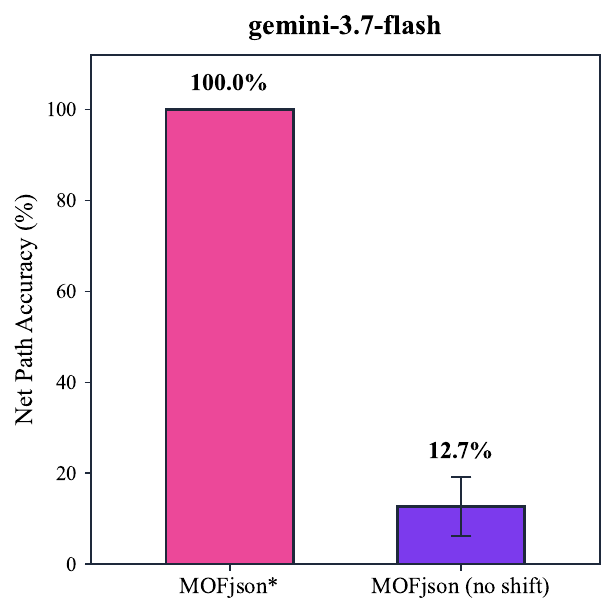}
\caption{\textbf{Impact of periodic shifts on Net path accuracy.} Performance of \texttt{gemini-3.7-flash} ($N=3$ runs). Zeroing edge shift vectors causes performance to drop from $100.0\%$ to $12.7\% \pm 6.4\%$.}
\label{fig:ablation-netpath}
\vspace{-10pt}
\end{wrapfigure}
\paragraph{Ablation.} To separate the periodic labels from the JSON formatting carrying them, we zeroed every edge's shift vector on Net path while grading against the unablated graph. Accuracy collapses from $100.0\%$ to $12.7\% \pm 6.4\%$ (Figure~\ref{fig:ablation-netpath}), below every baseline. We ablate \texttt{gemini-3.7-flash} alone because its 100.0\% baseline gives maximum headroom to isolate the drop.

\section{Discussion, Conclusion, and Future Work}
Making a MOF's periodic graph explicit pays off in proportion to how directly a question depends on the structure it exposes, and only once a model can use it (Section~\ref{sec:results}). The four models share neither training data nor architecture, yet the advantage tracks category difficulty consistently across them. A paired Wilcoxon signed-rank test confirms the ordering at $p<0.001$ in every category (Holm-Bonferroni, 12 comparisons), and Cliff's $\delta$ concentrates where topology is required (0.25--0.45 on Symmetry and Net path, $\approx$0.1 elsewhere). Pairs are clustered by structure across models, so the effective sample size is below $n$ and $p$-values are indicative (Table~\ref{tab:significance}, Appendix~\ref{tab:statsignifance}). The ablation (Section~\ref{sec:results}) locates the advantage in the periodic shift information rather than the format that contains it, although it only isolates the Net path questions.

\paragraph{Limitations.} Improvements depend on model capability: \texttt{apertus-v1.5-70b} benefits least, falling below baselines on Net path (Figure~\ref{fig:headline_netpath}, Appendix~\ref{sec:breakdownofaccuracy}). Token costs vary by serialization: topological queries (Connectivity, Net path, Chemical identity) include only the derived graph, taking $6\text{--}58\times$ fewer tokens than CIF ($\sim 200\text{--}1.7\text{k}$ vs.\ $\sim 11\text{k}$). Conversely, geometry queries (Symmetry) require the raw atomic layer, where MOFjson exceeds CIF ($\sim 13\text{k}$ vs.\ $\sim 11\text{k}$) (Figure~\ref{fig:token-counts}, Appendix~\ref{sec:token-analysis}).

MOFjson is a starting point: the ablation covers one mechanism on one model, and we report no chance-level baselines bounding structural reading versus recall. Beyond evaluation, explicit periodic graphs provide a foundation for spatial reasoning and generative design. Where language models have inherited representations built for database search, MOFjson argues for formats designed specifically for chemical reasoning.

\newpage
\section*{Acknowledgments}
This work was supported by the National Centre of Competence in Research (NCCR) Separations, funded by the Swiss National Science Foundation (SNSF). 

\section{Code and Reproducibility}
\label{sec:reproducibility}
All models were queried at temperature 0 with a fixed maximum output length, so run-to-run variance reflects the structure sample rather than decoding noise. No chain-of-thought or extended reasoning mode was enabled; every prompt requests a short, fixed-shape answer. Models were accessed via their providers' APIs, using the versioned identifiers listed in Section~\ref{sec:experiments}. Code and data are available at \url{https://github.com/schwallergroup/MOFJSON}. 
\bibliographystyle{unsrtnat} 
\bibliography{references}

\newpage
\appendix
\section{Appendix}
\subsection{HKUST-1 MOF: Schema, Structure, and Existing Identifiers}
\label{hkust-methodology}
\begin{figure}[H]
\centering
\includegraphics[width=0.85\textwidth]{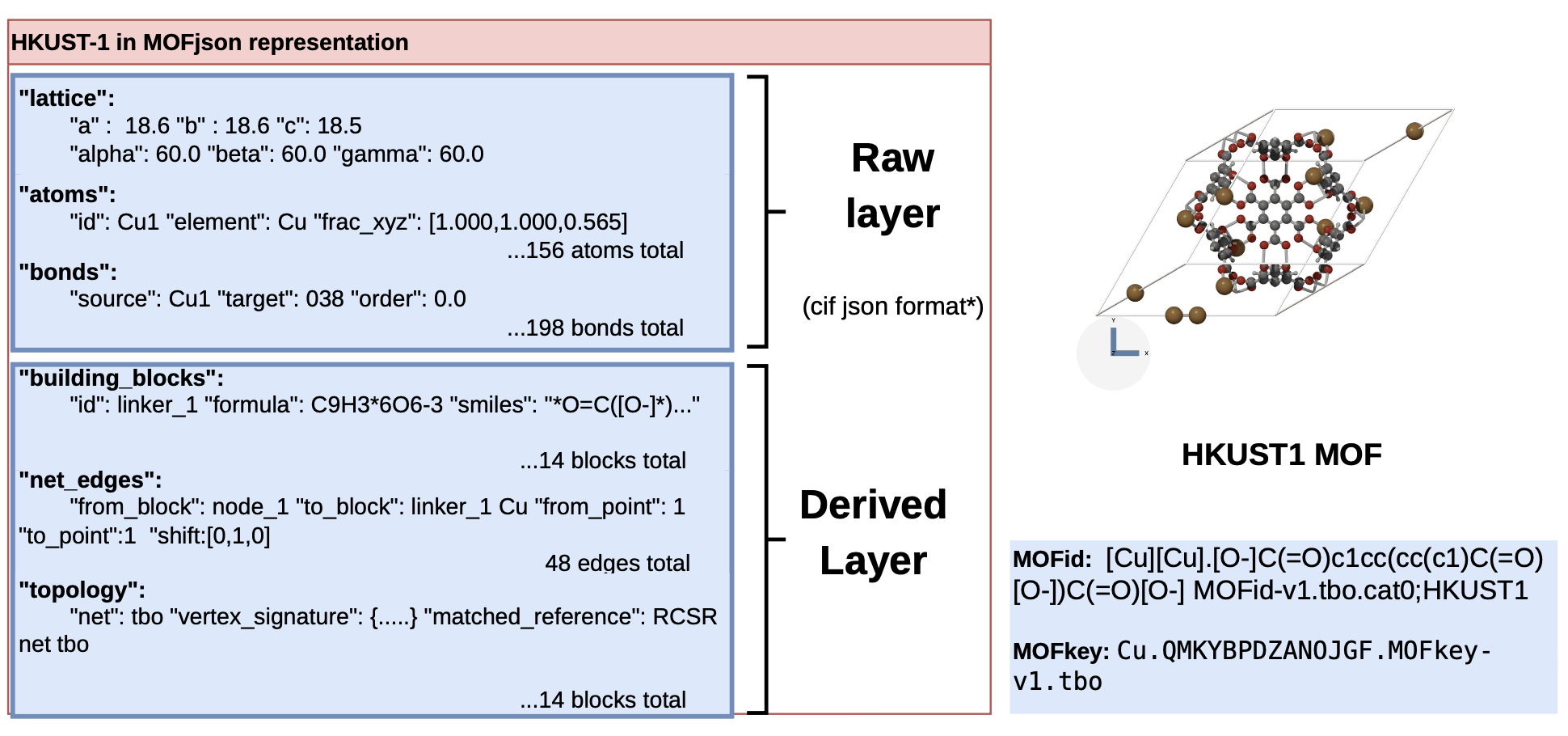}
\caption{\textbf{MOFjson makes HKUST-1's periodic graph explicit, where MOFid and MOFkey compress it into an opaque code.} Left: HKUST-1's MOFjson, showing the raw layer (lattice, atoms, bonds from CIF files) and the derived layer computed from it (building\_blocks, net\_edges, topology). Right: the same structure rendered in iRASPA software \cite{dubbeldam2018iraspa}, alongside its MOFid and MOFkey strings. Both identifiers encode HKUST-1's net symbol \texttt{tbo}.}
\label{fig:hkustmofjson}
\end{figure}

\subsection{Structural Fidelity of CIF to MOFjson Conversion}
\label{sec:conversion-fidelity}

\begin{figure}[H]
\centering
\includegraphics[width=0.62\textwidth]{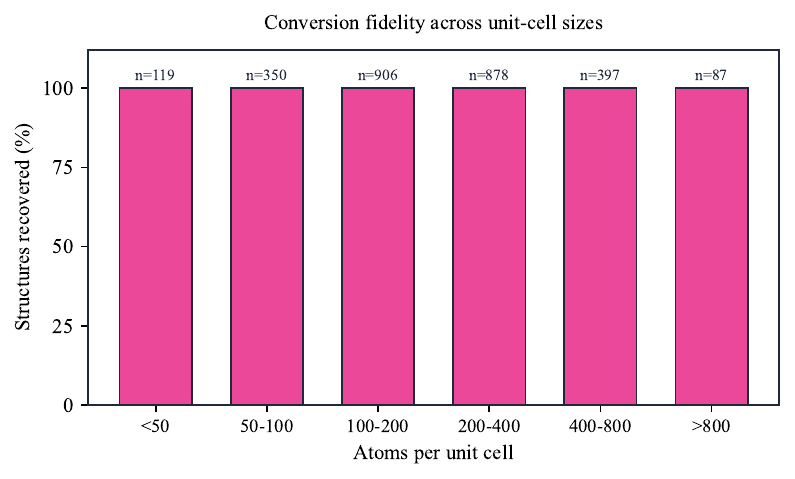}
\caption{\textbf{MOFjson conversion is lossless across the full CoRE MOF size range.} Each of the 2{,}737 computation-ready CIFs \cite{chung2014computation} is parsed with \texttt{pymatgen} \cite{pymatgen-ong2013python}, converted to the MOFjson \texttt{raw} layer, rebuilt into a structure from that layer alone, and compared against the source with \texttt{StructureMatcher}, which tests lattice parameters and site positions under symmetry operations rather than requiring byte-identical files. All 2{,}737 structures are recovered, from under 50 to over 800 atoms per unit cell ($n$ above each bar), so fidelity does not degrade with cell size. Only the \texttt{raw} layer is tested: the \texttt{derived} fields are computed from it and are not required for reconstruction, and a building-block partition is not uniquely invertible in any case.}
\label{fig:conversion-fidelity}
\end{figure}

\subsection{Best Representation by Model}
\label{tab:best-model-rep}
\begin{table}[H]
\centering
\begin{tabular}{lllr}
\toprule
Model & Best Representation & Question Category & Accuracy (\% $\pm$ SD) \\
\midrule
\texttt{claude-haiku-4.5} & CIF & Symmetry & 71.3 $\pm$ 7.6 \\
\texttt{gpt-5.6-luna} & MOFjson & Net path & 95.3 $\pm$ 4.2 \\
\texttt{gemini-3.7-flash} & MOFjson & Net path & 100.0 $\pm$ 0.0 \\
\texttt{apertus-v1.5-70b} & MOFjson & Chemical identity & 65.3 $\pm$ 8.1 \\
\bottomrule
\end{tabular}
\caption{Best-performing (representation, category) pair for each model, corresponding to the starred point in Figure~\ref{fig:strip-pair} (right). Accuracy is reported as mean $\pm$ SD across three runs.}
\label{tab:model-winners}
\end{table}

\subsection{Breakdown of Model Performance on the four Question categories}
\label{sec:breakdownofaccuracy}
\begin{figure}[H]
    \centering
    \includegraphics[width=\textwidth,height=0.33\textheight,keepaspectratio]{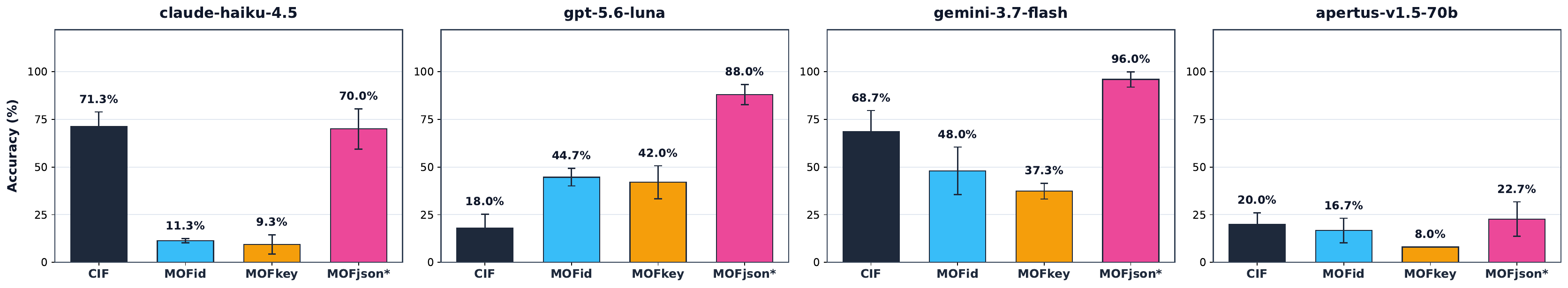}
    \caption{\textbf{MOFjson is consistently strong on symmetry questions.} It matches or leads CIF in three models, outperforming it in \textit{gpt-5.6-luna} (88.0\% vs.\ 18.0\%) and \textit{gemini-3.7-flash} (96.0\% vs.\ 68.7\%), while statistically tying in \textit{claude-haiku-4.5} (70.0\% vs.\ 71.3\%). Because MOFid and MOFkey lack lattice geometry, both consistently lag behind. The sole exception is \textit{apertus-v1.5-70b}, where performance across all representations collapses into a noisy 8--23\% band (22.7\% for MOFjson vs.\ 20.0\% for CIF). Thus, while MOFjson's lead over CIF depends on model capability, its edge over compressed string identifiers holds almost universally.}
    \label{fig:headline_symmetry}

    \vspace{0.8em}

    \includegraphics[width=\textwidth,height=0.33\textheight,keepaspectratio]{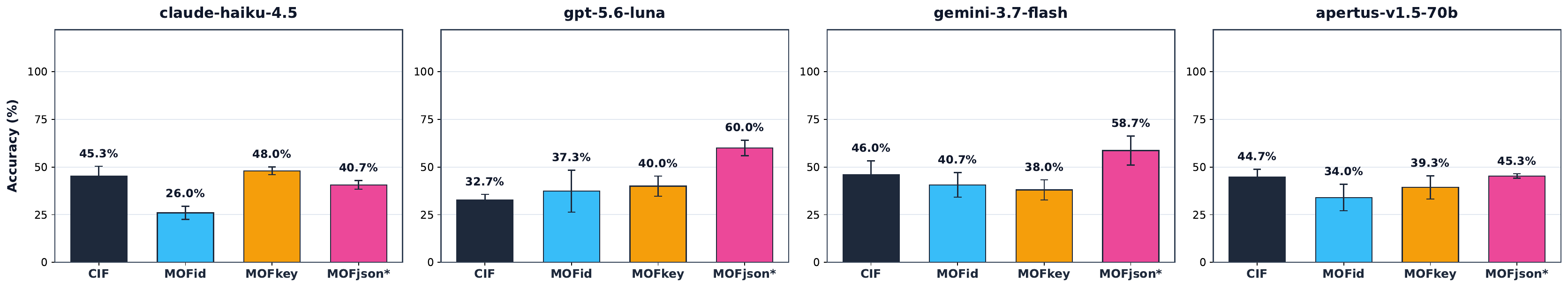}
    \caption{\textbf{MOFjson's connectivity advantage holds for most, but not all, models.} MOFjson leads clearly for \textit{gpt-5.6-luna} (60.0\% vs.\ 32.7--40.0\% for the baselines) and \textit{gemini-3.7-flash} (58.7\% vs.\ 38.0--46.0\%). For \textit{claude-haiku-4.5}, MOFkey scores highest (48.0\%) and MOFjson trails both CIF and MOFkey (40.7\%), and for \textit{apertus-v1.5-70b} all four representations sit within error of each other (34.0--45.3\%). Unlike symmetry, connectivity is not a category where MOFjson wins outright across models.}
    \label{fig:headline_connectivity}
\end{figure}

\begin{figure}[H]
    \centering
    \includegraphics[width=\textwidth,height=0.33\textheight,keepaspectratio]{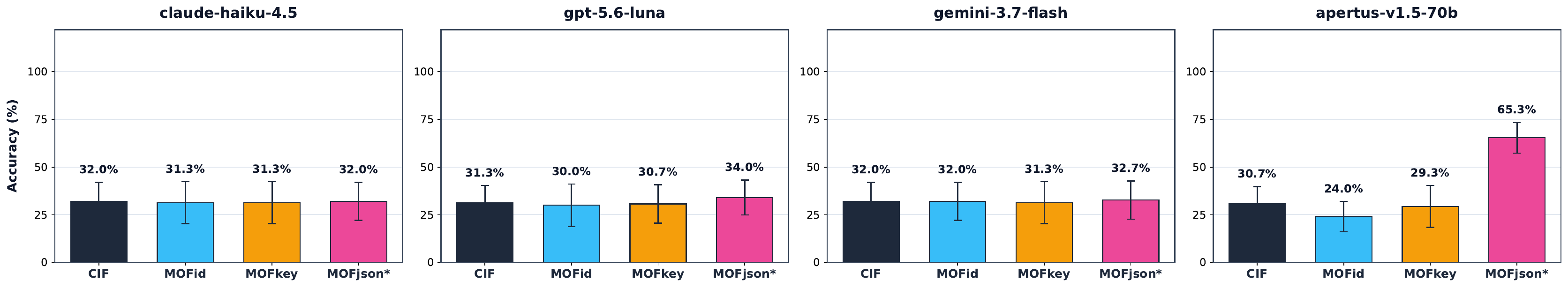}
    \caption{\textbf{Chemical identity shows no consistent MOFjson advantage.} For \textit{claude-haiku-4.5}, \textit{gpt-5.6-luna} and \textit{gemini-3.7-flash}, all four representations are statistically indistinguishable, clustered within 30--34\% regardless of format. Only \textit{apertus-v1.5-70b} separates, with MOFjson at 65.3\% against 24.0--30.7\% for the baselines. Because this appears in one model and nowhere else in the data results, we report it as a single-model observation rather than evidence of a representation effect, and treat Chemical identity as the category where the choice of representation matters least.}
    \label{fig:headline_identity}

    \vspace{0.8em}

    \includegraphics[width=\textwidth,height=0.33\textheight,keepaspectratio]{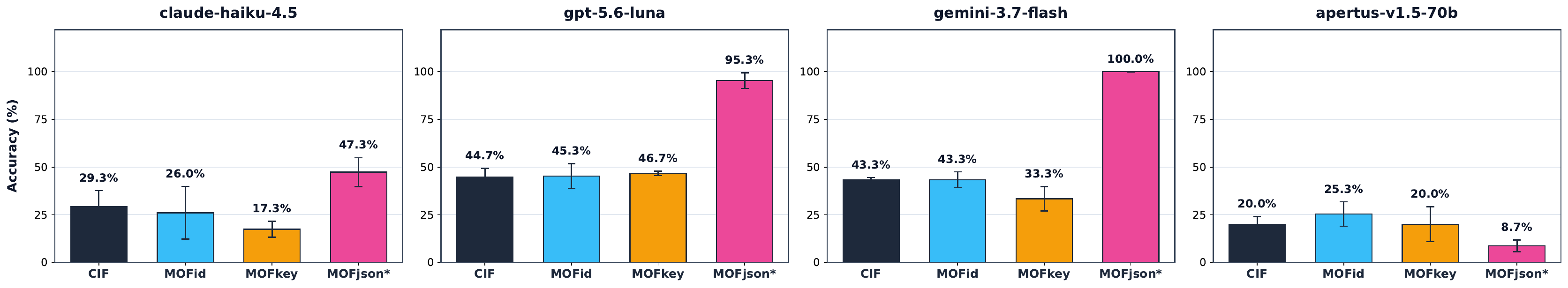}
    \caption{\textbf{Net path is where MOFjson's advantage is largest for most models.} MOFjson reaches 95.3\% for \textit{gpt-5.6-luna} and 100.0\% for \textit{gemini-3.7-flash}, against 33.3--46.7\% for every baseline in both cases, and leads more modestly for \textit{claude-haiku-4.5} (47.3\% vs.\ 17.3--29.3\%). For \textit{apertus-v1.5-70b} the result reverses: MOFjson scores 8.7\%, below all three baselines (20.0-25.3\%).}
    \label{fig:headline_netpath}
\end{figure}

\subsection{Paired Wilcoxon signed rank test and Cliff's Delta}
\label{tab:statsignifance}
\begin{table}[H]
\centering
\begin{tabular}{llrrr}
\toprule
Question Category & Baseline & $n$ & Cliff's $\delta$ & $p$ (Holm-adjusted) \\
\midrule
Symmetry & CIF & 588 & 0.25 & $<0.001$ \\
Symmetry & MOFid & 588 & 0.39 & $<0.001$ \\
Symmetry & MOFkey & 588 & 0.45 & $<0.001$ \\
Net path & CIF & 580 & 0.29 & $<0.001$ \\
Net path & MOFid & 580 & 0.28 & $<0.001$ \\
Net path & MOFkey & 580 & 0.34 & $<0.001$ \\
Connectivity & CIF & 588 & 0.09 & $<0.001$ \\
Connectivity & MOFid & 588 & 0.17 & $<0.001$ \\
Connectivity & MOFkey & 588 & 0.10 & $<0.001$ \\
Chemical identity & CIF & 592 & 0.10 & $<0.001$ \\
Chemical identity & MOFid & 592 & 0.12 & $<0.001$ \\
Chemical identity & MOFkey & 592 & 0.10 & $<0.001$ \\
\bottomrule
\end{tabular}
\caption{\textbf{Paired Wilcoxon signed-rank test comparing MOFjson against each baseline representation, per question category.} Every run draws 50 structures and renders each of them in all 4 representations, so a structure contributes one matched pair per model for each MOFjson-baseline comparison. The 3 runs draw independently, so a structure can recur across runs; de-duplicating the 150 draws leaves 145--148 distinct structures per category (2--5 repeats of the same MOF structure), and each contributes one paired observation per model, giving $n = 4 \times$ that count. Cliff's $\delta$ (range $-1$ to $1$; positive favours MOFjson) is the probability MOFjson scores higher on a given structure minus the probability it scores lower. $p$-values are Holm-Bonferroni corrected across all 12 tests. MOFjson's advantage is significant ($p<0.001$) against every baseline in every category.}
\label{tab:significance}
\end{table}

\section{Token Cost Analysis}
\label{sec:token-analysis}
\begin{figure}[H]
\centering
\includegraphics[width=0.8\textwidth]{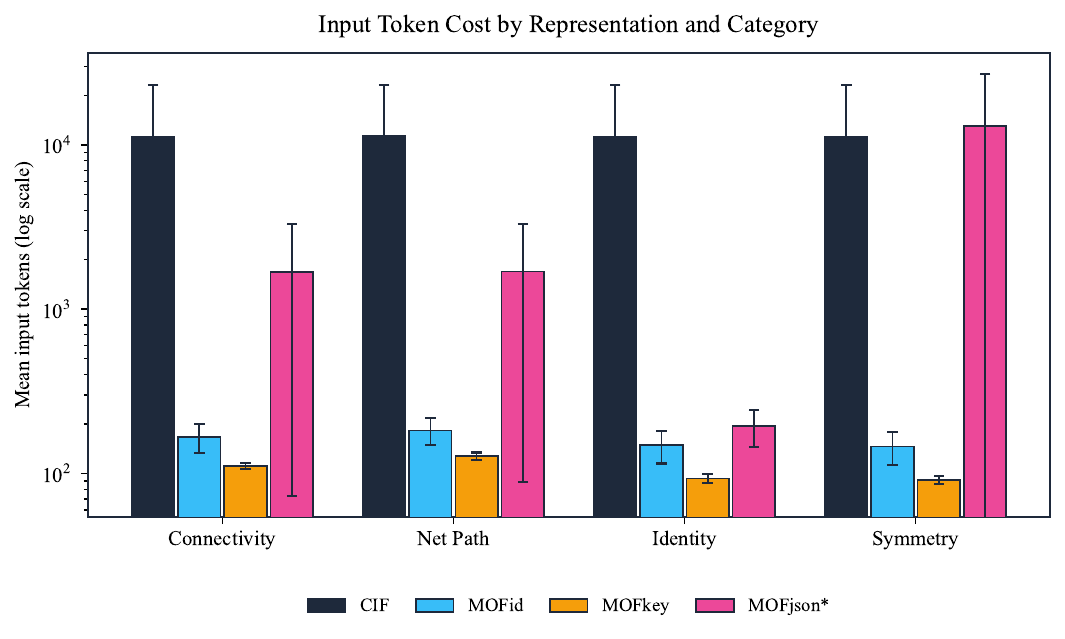}
\caption{\textbf{MOFjson's token cost is representation-efficient except for symmetry queries.} Measured in input tokens, MOFjson uses 6--58$\times$ fewer input tokens than CIF, approaching the compactness of MOFid and MOFkey. For symmetry, however, MOFjson requires the full per-atom coordinate list and exceeds even CIF's token cost, reflecting the overhead of its verbose JSON atom encoding relative to CIF's compact tabular atom-site format.}
\label{fig:token-counts}
\end{figure}

To assess the practical cost of each representation, we tokenize every prompt in our benchmark with the \texttt{cl100k\_base} encoding (via \texttt{tiktoken}) and report the mean input token count per representation, broken down by question category, in Figure~\ref{fig:token-counts}. Token counts are computed over the full question pool for each category, rather than the structure subsamples used in our accuracy evaluations, so that the reported costs reflect the benchmark as a whole rather than any single sampled run; error bars show one standard deviation across structures within each category.

\end{document}